\documentclass[10pt,twocolumn,letterpaper]{article}

\usepackage{cvpr}
\usepackage{times}
\usepackage{epsfig}
\usepackage{graphicx}
\usepackage{amsmath}
\usepackage{amssymb}

\usepackage{upgreek}
\makeatletter
\DeclareRobustCommand\onedot{\futurelet\@let@token\@onedot}
\def\@onedot{\ifx\@let@token.\else.\null\fi\xspace}
\def\eg{\emph{e.g}\onedot,~} 
\def\ie{\emph{i.e}\onedot,~} 
 
\def\etc{\emph{etc}\onedot} 
 
\def\etal{\emph{et al}\onedot}
\makeatother
\usepackage{url}
\usepackage{colortbl}
\usepackage[table]{xcolor}
\usepackage{multirow}
\usepackage{algorithmic}
\usepackage{algorithm}
\usepackage{array}
\usepackage{subfigure}
\usepackage{diagbox}
\usepackage{indentfirst}
\usepackage{ragged2e}
\usepackage{makecell}
\usepackage{xspace}
\usepackage{color}
\definecolor{mygray}{gray}{.9}
\definecolor{mypink}{rgb}{.99,.91,.95}
\definecolor{mycyan}{cmyk}{.1,0,0,0}
\usepackage{enumerate}
\usepackage{booktabs} 
\newcommand{\PreserveBackslash}[1]{\let\temp=\\#1\let\\=\temp}
\newcolumntype{C}[1]{>{\PreserveBackslash\centering}p{#1}}
\newcolumntype{R}[1]{>{\PreserveBackslash\raggedleft}p{#1}}
\newcolumntype{L}[1]{>{\PreserveBackslash\raggedright}p{#1}}

\usepackage[breaklinks=true,bookmarks=false]{hyperref}

\cvprfinalcopy 

\def\cvprPaperID{19} 

\ifcvprfinal\fi

\begin{document}

\title{Dissecting Person Re-identification from the Viewpoint of Viewpoint
}

\author{Xiaoxiao Sun~ ~ Liang Zheng\\
	 Australian National University \\
   {\tt\small ~~~xxsunzrt@gmail.com ~ liang.zheng@anu.edu.au}
}

\maketitle
\thispagestyle{empty}

\begin{abstract}
Variations in visual factors such as viewpoint, pose, illumination and background, are usually viewed as important challenges in person re-identification (re-ID).
In spite of acknowledging these factors to be influential, quantitative studies on how they affect a re-ID system are still lacking.
To derive insights in this scientific campaign, this paper makes an early attempt in studying a particular factor, viewpoint.
We narrow the viewpoint problem down to the pedestrian rotation angle to obtain focused conclusions.
In this regard, this paper makes two contributions to the community.
First, we introduce a large-scale synthetic data engine, PersonX.
Composed of hand-crafted 3D person models, the salient characteristic of this engine is ``controllable''.
That is, we are able to synthesize pedestrians by setting the visual variables to arbitrary values. 
Second, on the 3D data engine, we quantitatively analyze the influence of pedestrian rotation angle on re-ID accuracy.
Comprehensively, the person rotation angles are precisely customized from 0$^{\circ}$ to 360$^{\circ}$, allowing us to investigate its effect on the training, query, and gallery sets.
Extensive experiment helps us have a deeper understanding of the fundamental problems in person re-ID.
Our research also provides useful insights for dataset building and future practical usage, e.g., a person of a side view makes a better query.
\end{abstract}

\section{Introduction}
Viewpoint, pose of person, illumination, background and resolution are a few visual factors that are generally considered as influential problems in person re-identification (re-ID).
Currently, major endeavor is devoted to algorithm design to mitigate their impact on the recognition system.
Therefore, despite qualitatively acknowledging the factors as influential, it remains largely unknown how these factors affect the performance quantitatively. 

\begin{figure}
\begin{center}
	\includegraphics[width=0.95\linewidth]{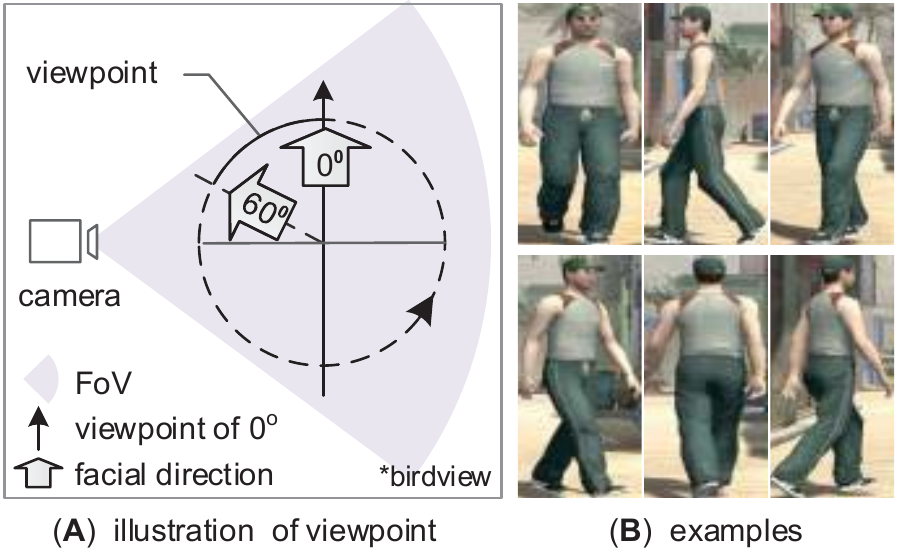}
\end{center}
\vspace{-1mm}
\caption{
	(\textbf{A}) Illustration of viewpoint of the birdview. 
	Viewpoint is defined as the rotation angle of a person relative to a manually defined degree of 0$^\circ$.
	The field of view (FoV) of a camera is shown. 
	(\textbf{B}) Examples of persons under different viewpoints.
}
\vspace{-3mm}
\label{fig:fig-vie}
\end{figure}

In this paper, we study one of the most important factors, \ie viewpoint.
Here, we denote viewpoint as the pedestrian rotation angle (Fig.~\ref{fig:fig-vie}). In what follows, we use viewpoint to replace pedestrian rotation angle unless specified. 
Since different views of a person contain different details, the viewpoint of a person influences the underlying visual data of an image, and thus the learning algorithm. Therefore, we aim to investigate the exact influence of viewpoint on the system.
This study will benefit the community from two aspects. 
(1) The conclusions of this research can guide for building the training set effectively. For example, finding that certain angles are more important for learning models of identifying pedestrians.
(2) It will advise for designing of query and gallery sets. By discovering viewpoints that are effective for re-ID accuracy, our research can potentially benefit the practical usage of re-ID systems.

In our attempt to reveal the influence of viewpoint, a notable obstacle is the lack of data.
Existing datasets might have a biased and fixed distribution of environmental factors.
In pedestrian viewpoint, for example, some angles might only have a few or even zero samples.
In another example, when studying illumination on a real-world dataset, conclusions are less convincing because the dataset might only has a specific illumination condition. 
Further, a fixed/static data distribution forbids us from exploring how the impact of viewpoint relates to other visual factors.
For example, the impact of viewpoint could be conditioned on the background, because background also affects feature learning.
To fully understand the role of viewpoint, we need to test its influence by changing the environment to either hard or easy modes. 
As such, without comprehensive and flexible data streams, we cannot make quantitative and scientific judgment of a visual factor's significance.

This paper makes two contributions to the community. 
First, we build a large-scale data synthesis engine named PersonX.
PersonX contains 1,266 manually designed identities and editable visual variables. It can simulate persons under various conditions. First of all, we demonstrate that existing re-ID models has \textbf{consistent accuracy trend} on both PersonX and real-world datasets \cite{zheng2015scalable,zheng2017unlabeled}. This observation suggests that \textbf{PersonX is indicative of the real world.} Moreover, as the name implies, the feature of PersonX is ``controllable''. Persons take controllable poses and viewpoints, and the environment is controlled \emph{w.r.t} the illumination, background, \emph{etc}.  
Persons move by running, walking \etc, under the controlled camera view and scene. We can obtain the exact person bounding boxes without external detection tools and thus avoid the influence of detection errors on the system.
Therefore, PersonX is indicative, flexible and extendable. It supports future research in not only algorithm design, but also scientific discoveries how environmental factors affect the system. 

Second, we dissect a person re-ID system by quantitatively understanding the role of person viewpoint. Three questions are considered.  (1) How does the \emph{viewpoint of the training set} influence the system?  (2) How does the \emph{query viewpoint} influence the retrieval? (3) How does the re-ID accuracy change under different \emph{viewpoint distributions of the testing set}?
To answer these questions, we perform rigorous quantification on pedestrian images regarding viewpoints. 
We customize the viewpoints of persons in the PersonX engine from 0$^{\circ}$ to 360$^{\circ}$. Both the control group and the experimental group are designed, so as to obtain convincing scientific conclusions. We also empirically study the real-world Market-1203 dataset where viewpoints of person are manually labeled. The empirical results are consistent with our findings on the synthetic data.

\section{Related Work}
We first review re-ID methods that improve the robustness against variations in pose, illumination, and background.
We then review methods based on synthetic data. 

\textbf{Against pose variance.} Some works \cite{farenzena2010person, zheng2017pose,cho2016improving,su2017pose,sarfraz2017pose} learn pose invariant representations for persons.
For example, Farenza \etal~\cite{farenzena2010person} utilizes body symmetry on the x-axis and asymmetry on the y-axis two axes to design a descriptor with pose invariance. 
Cho \etal~\cite{cho2016improving} quantize person poses into one of four canonical directions (front, right, back, left) to facilitate feature learning.
Zheng \etal~\cite{zheng2017pose} design the PoseBox to align different persons along the body parts.

\textbf{Against background variance.}
Some works reduce the influence of background \cite{bazzani2012multiple,xiao2017joint,chen2018person,song2018mask,tian2018eliminating,zhong2019invariance, zhong2019camstyle}.
For instance, Chen \etal~\cite{chen2018person} fuse the descriptors from the foreground person and the original image, such that the foreground is paid more attention to by the network. 
%
In \cite{song2018mask}, Song \etal use binary segmentation masks to separate foreground from the background. They then learn representations from the foreground and background regions, respectively. 
Zheng \etal~\cite{zheng2018pedestrian} apply STN to align pedestrian images, which reduces background noise and scale variances. 

\textbf{Against resolution variance.}
Resolution denotes the level of information granularity of an image. High resolution is typically preferred. But usually, the resolution level differs significantly across images. It thus affects the effectiveness of the learned features. 
To solve this problem, Jing \etal~\cite{jing2015super} design a mapping function that converts the features of low-resolution images into discriminative high-resolution features. 
alignment
In \cite{wang2018resource}, features from the bottom and top layers are concatenated during training and testing. Supervision signals are incorporated at each layer to train the multi-resolution features.  

\textbf{Against viewpoint variance.}
Learning viewpoint invariance is another focus \cite{gray2008viewpoint,wu2015viewpoint,bak2014improving,karanam2015person,zheng2017learning}. For example, Both \cite{gray2008viewpoint}\cite{karanam2015person} regard viewpoint variations as the most prominent problem. 
In this area, Gray \etal \cite{gray2008viewpoint} investigate the properties of localized features, while
Karanam \etal~\cite{karanam2015person} propose to learn dictionaries that can match person images captured under different viewpoints. 

\begin{figure*}[t]
\begin{center}
	\includegraphics[width=\linewidth]{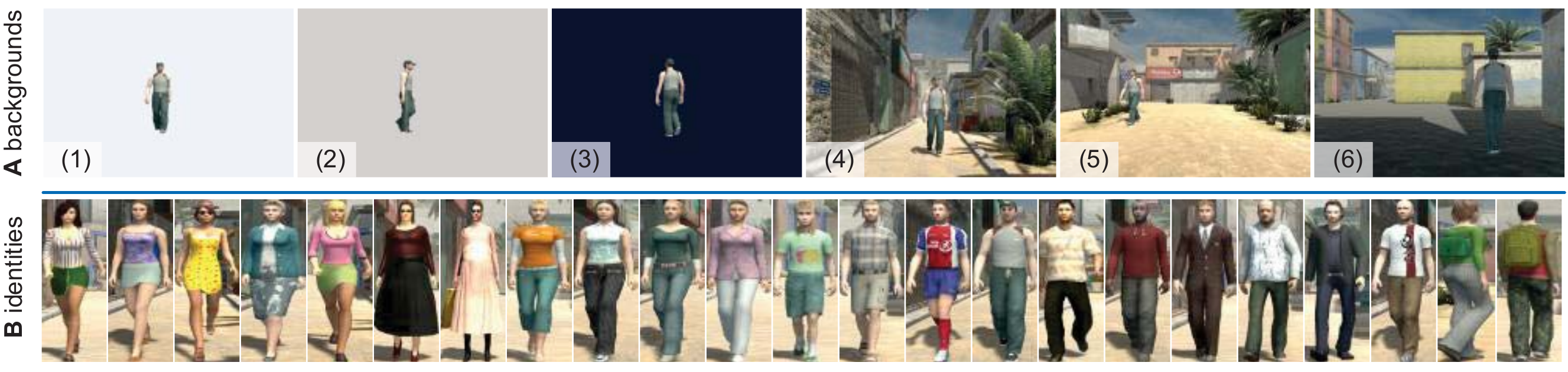}
\end{center}
\vspace{-3mm}
\caption{The PersonX dataset. 
	\textbf{A}: Background. In each background, a person can face toward a manually denoted direction, thus generating a controlled viewpoint.
	(1) - (3) represent backgrounds with uniform colors and (4) - (6) use street scenes as the background.
	\textbf{B}: Sample pedestrians bounding boxes in background (4). Various persons wearing various clothes are shown.
}
\label{fig:fig1}
\vspace{-3mm}
\end{figure*}

\textbf{Learning from synthetic data.} Leveraging synthetic data is a useful idea to alleviate the reliance on large-scale datasets. This strategy has been applied in problems like semantic segmentation~\cite{sankaranarayanan2018learning}, object tracking~\cite{gaidon2016virtual}, traffic vision research~\cite{li2018paralleleye} \etc
In the person re-ID domain, SOMAset \cite{barbosa2018looking} is a synthetic dataset with 50 person models and 11 types of outfits.
Barbosa \etal use SOMAset for training and test on real-world datasets. The accuracy was competitive. 
Bak \etal \cite{bak2018domain} also introduce a synthetic dataset SyRI including 100 characters. This dataset is featured by rich lighting conditions. A domain adaptation method is designed based on this dataset to fit real-world illumination distributions. 
Departing significantly from previous objectives of using synthetic dataset, this paper lays emphasis on quantitatively analyzing how visual factors influence the re-ID system. We derive useful insights by precisely controlling the simulator. This is a very early attempt of this kind in the community.  

\section{A Controllable Person Generation Engine}
\label{sec:PersonX}
\subsection{Description}
\textbf{Software.} The PersonX engine\footnote{The PersonX data engine, including pedestrian models, scene assets, project and script files \etc, are released at  \href{https://github.com/sxzrt/Dissecting-Person-Re-ID-from-the-Viewpoint-of-Viewpoint.git}{link}. } is built on Unity \cite{riccitiello2015john}.
We create a 3D controllable world containing 1,266 person models.
As a controllable system, it can satisfy various data requirements. 
In PersonX, the characters and objects look realistic, because the texture and materials of these models are mapped from the real world by scanning real people and objects.
The values of visual variables, \eg illumination, scenery and background, are designed to be editable. 
%
%
Therefore, PersonX is highly flexible and extendable.
%
%

\textbf{Identities.}
PersonX has 1,266 hand-crafted identities including $547$ females and $719$ males.
To ensure diversity, we hand-crafted the human models with different skin colors, ages, body forms (height and weight), hair styles, \etc
The clothes of these identities include jeans, pants, shorts, slacks, skirts, T-shirts, dress shirts, maxiskirt, \etc, and some of these identities have a backpack, shoulder bag, glasses or hat.
The materials of the clothes (color and texture) are mapped from images of real-world clothes.
The motion of these characters can be walking, running, idling (standing), having a dialogue \etc.
Therefore, the 3D models in PersonX look realistic. 
Figure~\ref{fig:fig1} (B) presents examples of identities of various ages, clothes, body shapes and poses.

\subsection{Visual Factors in PersonX}\label{sec:factors}
PersonX is featured by editable environmental factors such as illumination, cameras, backgrounds and viewpoints. Details of these factors are described below.

%

%
\textbf{Illumination.} Illumination can be directional light (sunlight), point light, spotlight, area light, \etc.
Parameters like color and intensity can be modified for each illumination type.
By editing the values of these terms, various kinds of illumination environment can be created.
\textbf{Camera.}
The configuration of cameras in PersonX is subject to different values of image resolution, projection, focal length, and height.

\begin{figure}[t]
\begin{center}
	\includegraphics[width=0.9\linewidth]{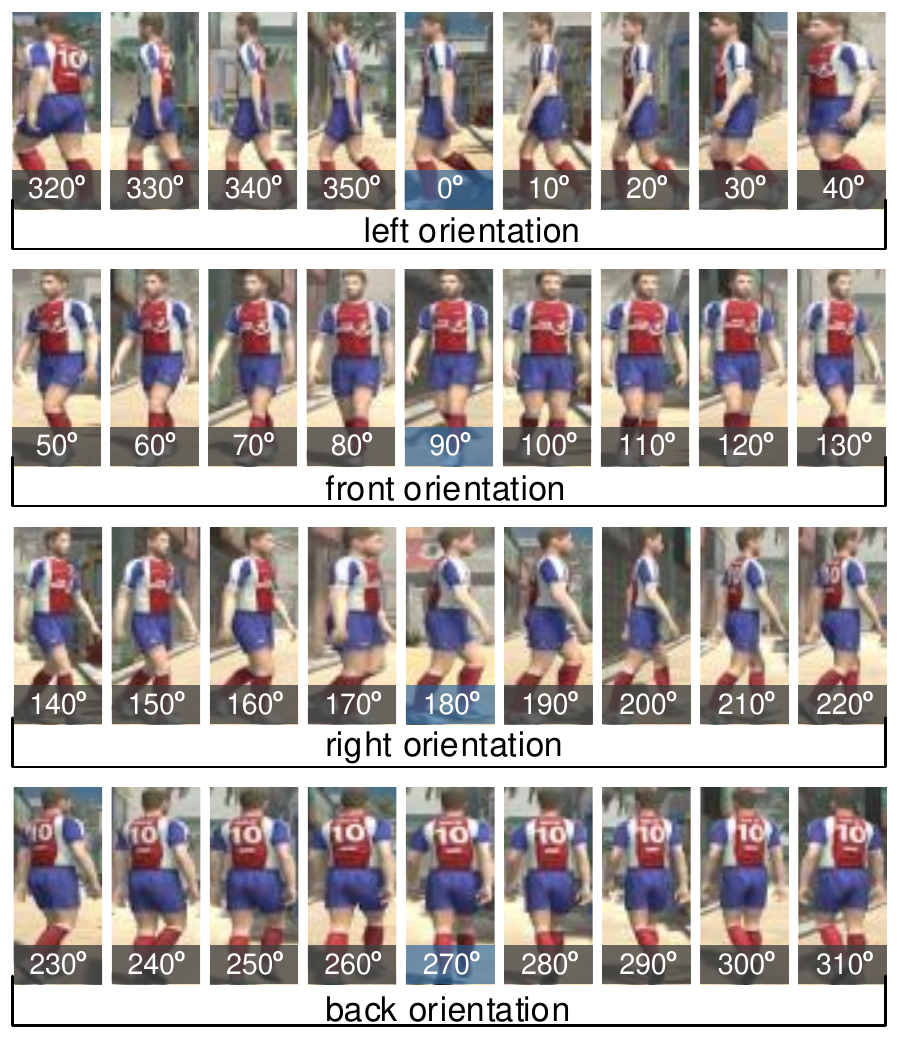}
\end{center}
\vspace{-3mm}
\caption{Definition of different viewpoints.
	Viewpoints of one identity are sampled at an interval of 10$^\circ$.
	Left orientation represents the set of the viewpoints that contains more information on the left side of the person, \ie 320$^\circ$ - 40$^\circ$.
	Similarly, other orientations of the pedestrian represent the sets of viewpoints containing front (50$^\circ$ - 130$^\circ$), right (140$^\circ$ - 220$^\circ$) and back (230$^\circ$ - 310$^\circ$) information.
	The viewpoints with \textcolor{blue}{the blue tags} represent the due left (0$^\circ$), front (90$^\circ$), right (180$^\circ$) and back (270$^\circ$) of a person.  
}
\label{fig:Rotation}
\vspace{-5mm}
\end{figure}

\textbf{Background.} 
Currently PersonX has six different backgrounds (Fig.~\ref{fig:fig1}). In each experiment, we set 2-3 different backgrounds/cameras views. In each background/camera view, a person moves freely in arbitrary directions, exhibiting arbitrary viewpoints relative to the camera.
%
%
In Fig.~\ref{fig:fig1}, backgrounds (4), (5) and (6) depict different street scenes.
Among the three scenes, backgrounds (4) and (5) share the same illumination and ground color, while background (6) is a shadowed region and the ground color is gray.
Meanwhile, backgrounds (1), (2) and (3) are pure colors and are used when background influence needs to be eliminated.
Because we simplify our system into two cameras, we use various combinations of these six cameras to create different re-ID environments. When not specified, all the cameras have a high resolution of 1024 $\times$ 768. 


%
%

\textbf{Viewpoint.}
Figure~\ref{fig:Rotation} presents image examples under specified viewpoints. Those images are sampled during normal walking.
Specifically, a person image is sampled every 10$^{\circ}$ from 0$^{\circ}$ to 350$^{\circ}$ (36 different viewpoints in total). Each viewpoint has 1 image, so each person has 36 images. The entire PersonX engine thus has 36 (viewpoints) $\times$ 1,266 (identities) $\times$ 6 (cameras) $=$ 
273,456 images.
%
For each person, the 36 viewpoints are divided into 4 groups, representing four orientations: left, front, right and back. We use ``left'' and ``due left'' to represent images with viewpoints from 320$^{\circ}$ to 40$^{\circ}$ (\ie left orientation), the image of 0$^{\circ}$, respectively. This convention applies for other orientations.
%
%

Comparisons of PersonX and some existing re-ID datasets are presented in Table \ref{table:Datasets}.
%
There are two existing synthetic datasets, SyRI \cite{bak2018domain} and SOMAset \cite{barbosa2018looking}. SyRI is used as an alternative data source for domain adaptation and does not have the concept of cameras.  SOMAset contains 250 cameras, which are uniformly distributed along a hemisphere around each person. Neither datasets are freely editable by the public.
In comparison, PersonX has configurable backgrounds and much more identities. Importantly, it can be edited/extended not only for this study, but also for future research in this area. 
%
%
%
%


\section{Benchmarking and Dataset Validation}
\label{sec:validation}
In this section, we aim to validate that PersonX is indicative of the real world, such that conclusions derived from this dataset can be of value to practice.  
\begin{table}[t]\footnotesize
\begin{center}
	\setlength{\tabcolsep}{2.1mm}{
		\begin{tabular}{l|l|c|r|c|c} 
			\Xhline{1.2pt}
			\multicolumn{2}{c|}{dataset}	&  \#identity & \multicolumn{1}{c|}{\#box} & \#cam. &  \footnotesize view\\
			\hline 
			\multirow{5}{*}{\rotatebox{90}{real data}}	& Market-1501~\cite{zheng2015scalable} &1,501  &32,668 &6 & N \\
			&Market-1203~\cite{zheng2015scalable} & 1,203 &8,569 & 2 & Y \\
			&MARS~\cite{zheng2016mars} & 1,261 & 1,191,003  & 6  & N \\
			&CUHK03~\cite{li2014deepreid} & 1,467&   14,096 & 2 & N\\
			&Duke~\cite{ristani2016performance}  &  1,404&  36,411  & 8 & N \\
			\hline
			\multirow{6}{*}{\rotatebox{90}{synthetic data}}	&SOMAset~\cite{barbosa2018looking}  & 50 & 100,000& 250 & N\\
			& SyRI~\cite{bak2018domain}  & 100 & 1,680,000   & -- & N\\
			&PersonX & 1,266 & 273,456 & 6 & Y \\
			&PersonX$_{123,456}$ & 1,266 & 136,728 & 3 & Y \\
			&PersonX$_{12, 13}$& 1,266 & 91,152   & 2 & Y \\
			&PersonX$_{45, 46}$ & 1,266 & 91,152 & 2 & Y \\
			\Xhline{1.2pt} 			
	\end{tabular}}
	\vspace{1em}
	\caption{Comparison of real-world and synthetic re-ID datasets.
		``View'' denotes whether the dataset has viewpoint labels.
		\label{table:Datasets}}
\end{center}
\vspace{-2.5em}
\end{table}

\subsection{Methods and Subsets}
We use IDE+ \cite{zhong2018camera}, triplet feature \cite{hermans2017defense} and PCB \cite{sun2018beyond} for our purpose.
IDE+ is implemented on ResNet50.
During training, the batch size is set to 64 and the model is trained for 50 epochs. The learning rate is initialized to 0.1 and decays to 0.01 after 40 epochs.
The model parameters are initialized with the model pre-trained on  ImageNet.
For triplet feature, the number of identities per batch is set to 32 and number of images per identity is set to 4. So the batch size is 32 $\times$ 4 $=$ 128.
The learning rate is initialized to $2\times10^{-4}$ and decays after 150 epochs (300 epochs in total).
Training of PCB follows the standard setup described in \cite{sun2018beyond}.

Through combinations of the six backgrounds described in Section~\ref{sec:factors}, PersonX has the following subsets. 

\begin{itemize}
    \vspace{-2mm}
    \item PersonX$_{12}$. It has backgrounds (1) and (2). Both are pure color backgrounds; the colors are similar.
    \vspace{-2mm}
    \item PersonX$_{13}$. The two cameras face backgrounds (1) and (3). The color difference between the two backgrounds is significant than that in PersonX$_{12}$.
    \vspace{-2mm}
    \item PersonX$_{123}$. This is a three-camera system, comprising backgrounds (1), (2) and (3).
    \vspace{-2mm}
    \item PersonX$_{45}$. It contains backgrounds (4) and (5) of street scenes. The two backgrounds are close in scene and illumination.
    \vspace{-2mm}
    \item PersonX$_{46}$. It consists of backgrounds (4) and (6). The two backgrounds have larger disparity than PersonX$_{45}$. 
     \vspace{-2em}
    \item PersonX$_{456}$. It is a three-camera system consisting of backgrounds (4), (5) and (6). 
    \vspace{-2mm}
\end{itemize}

Overall, PersonX$_{12}$,  PersonX$_{13}$ and PersonX$_{123}$ are simple subsets, while PersonX$_{45}$, PersonX$_{46}$ and PersonX$_{456}$ are more complex ones. Moreover, we introduce low-resolution subsets to create more challenging settings. We edit the image resolution of PersonX$_{45}$, PersonX$_{46}$ and PersonX$_{456}$ from 1024 $\times$ 768 to 512 $\times$ 242 (for images of FoV). We use ``-lr'' to represent low-resolution subsets. 
%

\begin{figure*}[tp]
	\begin{center}
		\includegraphics[width=\linewidth]{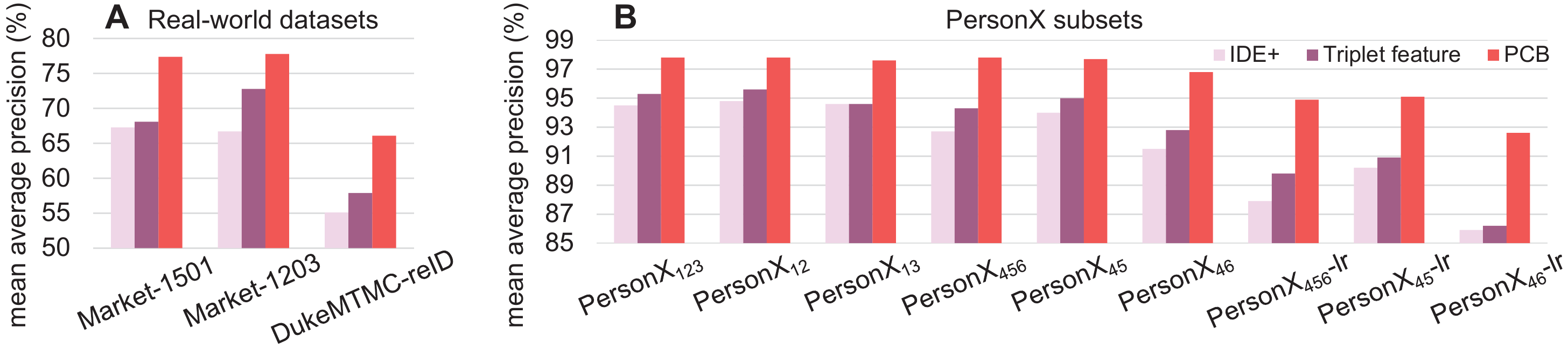}
	\end{center}
	\vspace{-4mm}
	\caption{Re-ID mean average accuracy (mAP, \%) of IDE+, triplet feature, and PCB on \textbf{(A)} real-world datasets and \textbf{(B)} the PersonX subsets. 
		``lr'' means the frames are low resolution of 512 $\times$ 242 instead of the original resolution 1024 $\times$ 768. 
	}
	\label{fig:benchmark}
	\vspace{-3mm}
\end{figure*}

For benchmarking, we randomly sample 410 identities for training and the rest 856 identities for testing.
In each camera, an identity has 36 images, \ie 36 viewpoints, from which one image is selected as the query during testing. Therefore, the three-camera subsets, \ie PersonX$_{456}$ and PersonX$_{123}$, contain 44,280 (410 $\times$ 36 $\times$ 3) training and 92,448 (856 $\times$ 36 $\times$ 3) testing images. The two-camera subsets have 29,520 (410 $\times$ 36 $\times$ 2) training and 61,632 (856 $\times$ 36 $\times$ 2) testing images.

\subsection{System Validation}
\label{sec:data-val}
We evaluate the three algorithms on both real-world and synthetic datasets. We use the standard evaluation protocols \cite{zheng2015scalable,zheng2017unlabeled}. Results are reported in Fig \ref{fig:benchmark}. We observe three characteristics of PersonX.

First, \textbf{eligibility}. We find the performance trend of the three algorithms is similar between PersonX and real-world datasets. 
On Market-1501 and DukeMTMC, for example, PCB has the best accuracy, and the performance of IDE+ and triplet feature is close. That is, PCB $\succ$ triplet $\approx$ IDE+. This is consistent with findings in \cite{sun2018beyond}. On the synthetic PersonX subsets, the performance trend is similar: IDE+ and triplet feature have similar accuracy; PCB is usually 2\%-3\% higher than them. These observations suggest that PersonX is indicative of the real-world and that future conclusions derived from PersonX can be of real-world value.

Second, \textbf{purity}. The re-ID accuracy on PersonX subsets (Fig.~\ref{fig:benchmark} (B)) are relatively high compared to the real-world datasets (Fig.~\ref{fig:benchmark} (A)).
It does not mean these subsets are ``easy''.
In fact, the high accuracy is what we design for, as it excludes the influence of the environmental factors.
In other words, these subsets are \emph{oracle}:  images are high-resolution, and the scenes have normal sunlight and relatively consistent backgrounds. These subsets are thus ideal ones for studying the impact of viewpoints. 

Third, \textbf{sensitivity}. We show that these subsets are sensitive to the changes in the environment.
For example, background variation in PersonX$_{46}$ is much larger than PersonX$_{45}$. As such, we observe that mAP in PersonX$_{46}$ is lower by 1\% - 3\% for different algorithms. 
Similarly, the background in PersonX$_{12}$ is much simpler than  PersonX$_{46}$, which causes mAP on PersonX$_{46}$ to be lower than on PersonX$_{12}$. 
Further, when these subsets are manually edited to be low resolution, we observe a significant mAP drop.
For example, the mAP drop from PersonX$_{46}$ to PersonX$_{46}$-lr is about 6\%.
The above comparisons demonstrate that PersonX subsets are sensitive to background complexity, variation between cameras, and image resolution. This is consistent with our intuition and indicates that PersonX is useful in studying the influence of visual factors. 

The above discussions indicate that PersonX is indicative of the real-world trend, has strictly controlled environment variables, and is reasonably sensitive to environmental changes.
We believe PersonX will be a useful tool for the community and encourage the development of robust algorithms and scientific analysis.

\begin{figure*}[tp]
	\begin{center}
		\includegraphics[width=\linewidth]{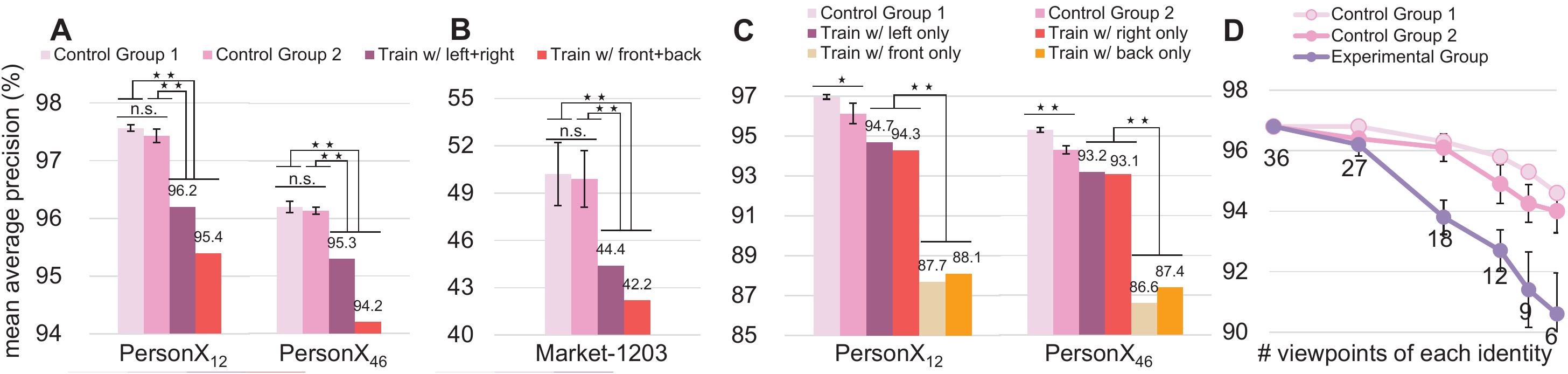}
	\end{center}
	\vspace{-2mm}
	\caption{Re-ID accuracy (mAP, \%) when the training set has missing orientations/viewpoints. \textbf{A} and \textbf{B}: we use two orientations for training. For example, we can train with left and right orientations only (see Fig. \ref{fig:Rotation} for the definition of orientation). \textbf{C}: we train with one orientation \emph{only}, \ie left, right, front, or back orientation. For each dataset, we have two control groups. 
	\textbf{D}: Impact of missing continuous viewpoints on PersonX$_{46}$.
	The horizontal axis is the remaining number of viewpoints and vertical axis is the mAP.
	In the experimental group, continuous viewpoints are removed. The number on this curve denotes the remaining number of viewpoints. ``n.s." represents that the difference between results is \textbf{not statistically significant} (\ie $p-\text{value}> 0.05$). $\star$ corresponds to \textbf{statistically significant} (\ie $0.01< p-\text{value}< 0.05$). $\star\star$ means the difference between results is  \textbf{statistically very significant} (\ie $0.001< p-\text{value}< 0.01$). 
	}
	\label{fig:train1}
		\vspace{-3mm}
\end{figure*}


\section{Evaluation of Viewpoint}
We evaluate the impact of viewpoint on person re-ID.
The experiment is based on PCB \cite{sun2018beyond}. We note that other standard re-ID methods (\eg IDE+) can draw similar conclusions. 
%
%
Three questions will be investigated in the following subsections: how does the viewpoint in (1) the training set, (2) the query set, and (3) the gallery set affect the re-ID accuracy? For  clearer understanding, we mainly show figures in this section. Detailed numbers are provided in the supplementary material.
%
%
%

\subsection{How Do Viewpoint Distributions in the Training Set Affect Model Learning?}\label{sec:train_view}

\textbf{Experiment design.}  Initially, the subsets contain all the viewpoints for the training and testing IDs. That is, a person has 36 images under each camera.
In this section, to study the influence of missing viewpoints in the training set, 
we remove specific orientations from the training set. The orientations refer to left, front, right and back shown in Fig.~\ref{fig:Rotation}.
We design the following training sets.
\begin{itemize}
      \vspace{-2mm}
    \item Control group 1. We \emph{randomly} select half (18 out of 36) or a quarter (9 out of 36) images of each identity for training.
    \vspace{-2mm}
    \item Control group 2. The training set is constituted by \emph{randomly} selecting half (18 out of 36) or a quarter (9 out of 36) viewpoints for each identity.
    \vspace{-2mm}
    \item Experimental group 1. Train with two orientations. The training images exhibit two orientations, left+right or front+back. The training set is thus half of the original training set. 
    \vspace{-2mm}
    \item Experimental group 2. Train with one orientation. The training set has one orientation, \ie left, right, front, or back. The training set becomes a quarter of the size of the original training set. 
  \vspace{-2mm}
\end{itemize}

\textbf{Discussions.} The experimental groups are used to assess the impact of missing viewpoints in the training set. To cancel out the result influence of reduced training images and the non-uniform viewpoint distributions of the experimental groups, we further design two control groups, where the number of images used for training is the same with the above two experimental groups. The first control group removes images randomly, and the second control group removes viewpoints randomly. For control group 1 and 2, we repeat our experiment 5 times and report the average re-ID accuracy. Using the two control groups, we highlight the impact of the missing viewpoints in the training set.
 
%
%
%

\textbf{Result analysis.} Using the experimental groups and control groups designed above, we summarize the key experimental results in Fig.~\ref{fig:train1}. These results are reported on two synthetic datasets, PersonX$_{12}$ and PersonX$_{46}$, and a real-world dataset, Market-1203. We mainly use the mean average precision (mAP) for evaluation, as it provides a comprehensive assessment of the system's ability to retrieve all the relevant images. From these results, we have several observations as follows. 

First, the two control training sets have similar accuracy, and control group 2 is slightly inferior. Control groups 2 has some specific viewpoint missing, while control group 1 has images randomly missing. This indicates that viewpoint comprehensiveness is important for a training set. 

Second, removing continuous viewpoints in the training set causes more accuracy drop than removing random viewpoints or random images. In control group 2, the viewpoints are randomly removed. In the two experimental groups, continuous viewpoints are removed. The inferior accuracy of the experimental groups (see Fig. \ref{fig:train1} (A), (B), (C) and (D)) indicate that continuous viewpoints are more important. Further, Fig. \ref{fig:train1} (D) demonstrates an increasing performance gap as more viewpoints are removed. This observation is intuitive because continuous viewpoints encode appearance cues that cannot be recovered by other viewpoints and once lost, will cause system degradation.

Third, from Fig. \ref{fig:train1} (A) and (B), when the training set only has two orientations (left+right or front+back), we observe a significant accuracy drop compared with the control groups. 
Similarly, Fig. \ref{fig:train1} (C) indicates that a training set with only one orientation also deteriorates the re-ID accuracy when compared with the control groups. 

Fourth and importantly, left/right orientations make a better training set than front/back orientations. From Fig. \ref{fig:train1} (A), (B) and (C), when the training set is composed of left/right orientations, the re-ID accuracy is higher than training sets with front/back orientations.
For example, when using a training set composed of ``front+back'' orientations, the mAP score in Fig.~\ref{fig:train1} (A) is 0.8\%-1.1\% lower than a model trained with  ``left+right'' orientations. 
On the real-world dataset, Market-1203 in Fig.~\ref{fig:train1} (B), 
the mAP of a model trained with left and right orientations is 2.2$\%$ higher than learning from the viewpoint from front and back. It indicates that data synthesis is indicative of the real world to some extent. Similarly, when the training set only has one orientation (Fig.~\ref{fig:train1} (C)), the left or right orientations are significantly more beneficial than the front or back orientations. The mAP gap can be as large as 6\%. Note that for evaluating training sets with only one orientation, we do not use Market-1203. This is because Market-1203 does not have sufficient training samples under each orientation.

Regarding the observation that left/right orientations are more useful than front/back orientations, we provide a plausible reason below. 
For pedestrians, the left or right orientations reflect important general information, such as color, outfit (\eg {long or short sleeve, pants, shorts}) \etc. In comparison, the front and back views capture more detailed appearance cues such as prints of clothes, face, \etc. 
As such, a model trained with left/right viewpoints encodes the general appearance knowledge about pedestrians; a model trained with front/back viewpoints somehow has abilities that are useful in recognizing specifically the front/back views but might be abundant for the side views. 
In other words, if a true match is of the left of right orientation that does not present as much texture details, a model trained with front or back orientations may not work well.
On the other hand, a model trained with left or right viewpoints is good at recognizing the clothes appearance, so its performance will not deteriorate much when identifying pedestrians under front or back orientations.

\begin{figure}[t]
\begin{center}
	\includegraphics[width=1\linewidth]{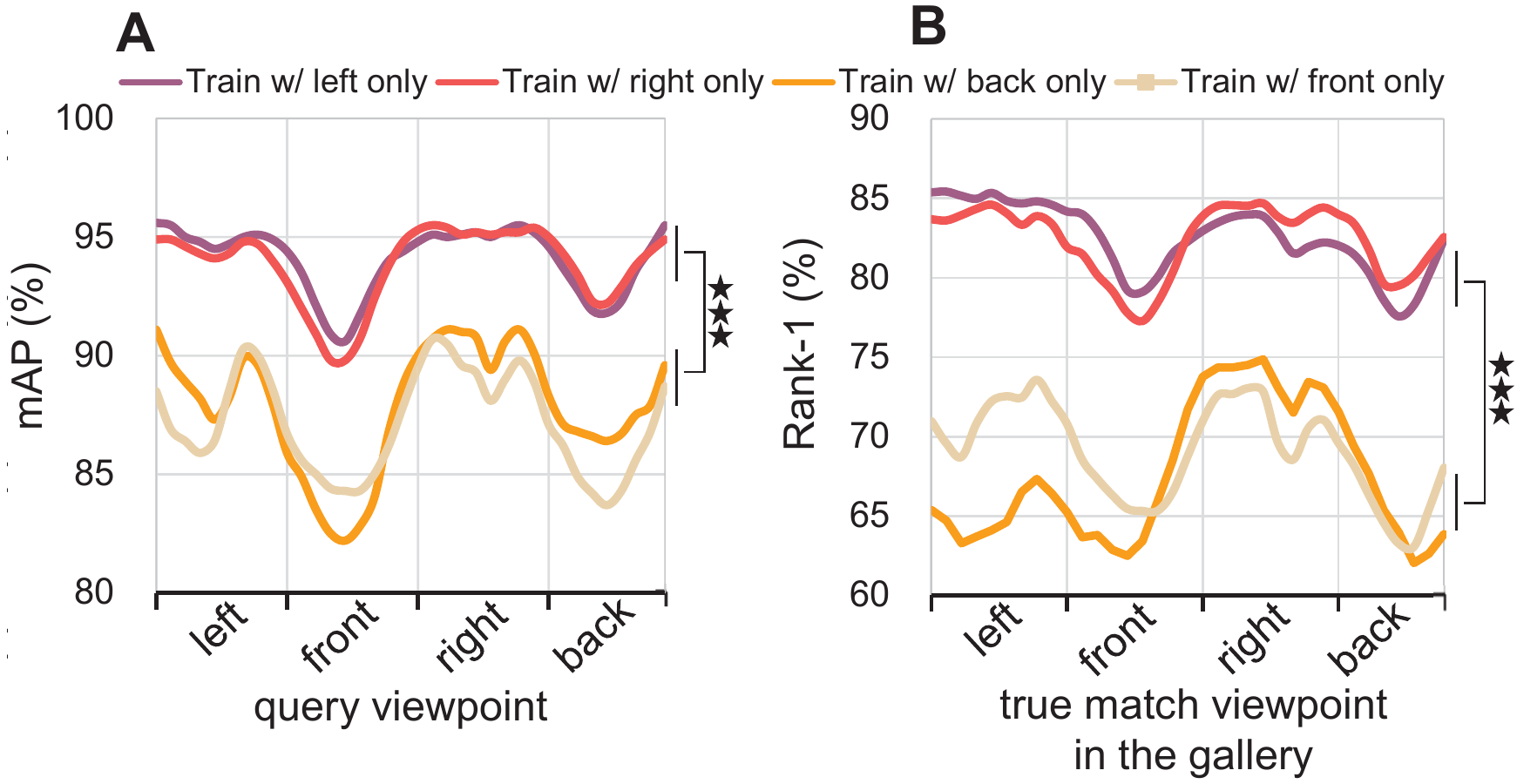}
\end{center}
\vspace{-3mm}
\caption{Evaluation of models trained on one orientation only.
\textbf{A:} query viewpoint change vs. mAP. Query viewpoint changes from left, front, right to back. True matches in the gallery are uniformly distributed. Since a query has multiple true matches, we use mAP to measure accuracy.
 \textbf{B:} true match viewpoint vs. rank-1. The true match viewpoint changes from left, front, right to back. 
    Query viewpoints are uniformly distributed. \emph{Each query has only one true match in the gallery}, so the rank-1 accuracy is used for evaluation. 
    $\star\star\star$ means that the difference between results of models trained on left/right and front/back orientations is \textbf{statistically extremely significant} (\ie $ p-\text{value}< 0.001$).}
\label{fig:training-query}
\vspace{-3mm}
\end{figure}

\textbf{A further study.} To further understand the superiority of left/right orientations in training, we quantify the query viewpoint and the true match viewpoint into four orientations, too. Results are shown in Fig. \ref{fig:training-query}. Here, training sets are constructed with only one orientation. 
First, when the query images exhibit only one orientation, and when the true match viewpoint distributes uniformly in the gallery, a training set with left or right orientations is superior to that with front or back orientations.
Second, we assume a single viewpoint for all the true matches in the gallery. 
We also assume a single true match for each query. Four models are trained with images solely from one of the four orientations. We show the rank-1 accuracy of the four models in Fig. \ref{fig:training-query} (B). Note that the viewpoint distribution in query set is uniform. 
In our observation, for a true match to be retrieved, using models trained on the left or right orientations yields higher accuracy than models trained on front or back orientations.
Therefore, regardless of the viewpoint distribution in the gallery or query set, a person re-ID model trained with left or right orientations performs favorably than trained with  front or back orientations. 
\vspace{2mm}
\noindent	\fbox{\parbox{0.46\textwidth}{	\textbf{Subsection conclusions}
	\vspace{-2mm}
	\begin{itemize}
		\item  Missing viewpoint compromises training.
		\vspace{-2mm}
		\item Missing continuous viewpoints are more detrimental than missing randomly viewpoints. 
		\vspace{-2mm}
		\item  When limited training viewpoints are available, left/right orientations allow models to be better trained than front/back orientations.
  \vspace{-2mm}
\end{itemize}}}



\begin{figure}[t]
\begin{center}
	\includegraphics[width=.85\linewidth]{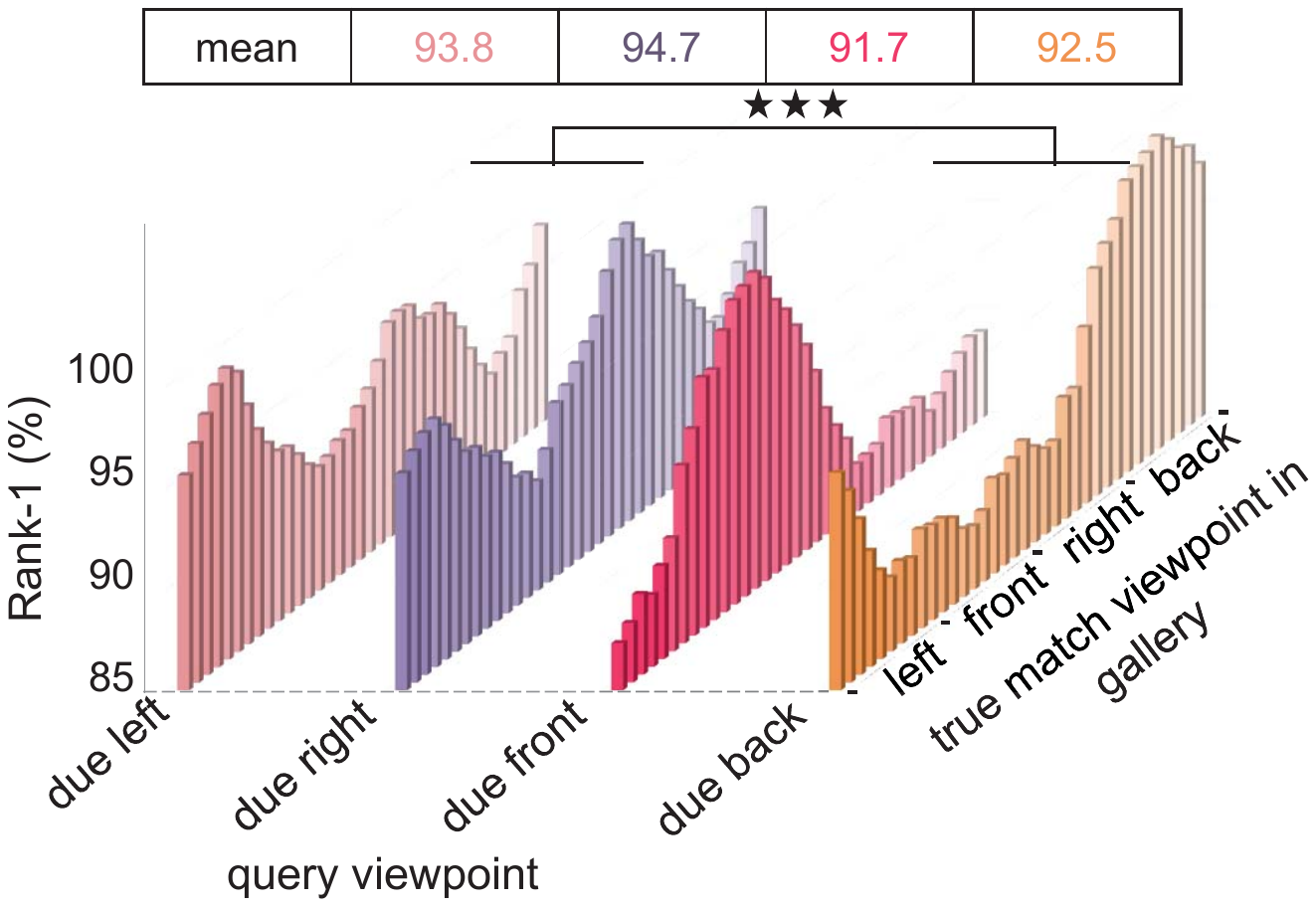}
\end{center}
\vspace{-2mm}
\caption{Impact of query viewpoint on system performance on PersonX$_{45}$. Four viewpoints are evaluated, \ie due left (0$^\circ$), right (180$^\circ$), front (90$^\circ$) and back (270$^\circ$). In the gallery, there is only one true match for each query.
The true match viewpoint varies from 0$^\circ$ to 350$^\circ$ (deep axis).
Under each query viewpoint, we report 36 rank-1 scores obtained by the query to retrieve 36 types of true match viewpoints.
$\star\star\star$ means the difference between retrieval results of due left/right and due front/back is \textbf{statistically extremely significant} (\ie $ p-\text{value}< 0.001$).
On the top, we show the average rank-1 accuracy for each query viewpoint.}
\vspace{-3mm}
\label{fig:singel-rotation}
\end{figure}

\subsection{How Does Query Viewpoint Affect Retrieval?}
We study how query viewpoint influences re-ID results.

\textbf{Experiment design.}
We train a  model on the original training set comprised of every viewpoint. We modify the query viewpoints to see its effect during testing.
Specifically, the viewpoint of a probe image can be set to the \emph{due left} (0$^\circ$),  \emph{due front} (90$^\circ$),  \emph{due right} (180$^\circ$) or \emph{due back} (270$^\circ$) to represent different sides of person.
During retrieval, we assume only one true match in gallery; the true match contains the same person as the query, and its viewpoint is between 0$^\circ$ and 350$^\circ$. Viewpoints of the distractor gallery images are images of all other persons.

\textbf{Result analysis.}
Figure~\ref{fig:singel-rotation} presents the results obtained by the above query and gallery images. We use PersonX$_{45}$ for training and testing. We have several observations. 

First, when the viewpoint of the true match is similar to the query, the highest re-ID accuracy can be achieved. 
For example, the maximum rank-1 values of due left queries correspond exactly to the due left true match in the gallery. Under the same viewpoint, the query and true match are different only in illumination and background. This indicates that viewpoint differences between two to-be-matched images cause performance drop. 

Meanwhile, queries of the due left and the due right viewpoint lead to a higher average rank-1 accuracy than queries of due front and due back viewpoints. For example, the accuracy of the due left queries and the due front queries is 93.8\% and 91.7\%, respectively. 
It is noteworthy that in Section \ref{sec:train_view} and Fig. \ref{fig:training-query}, we can have a similar observation regarding the superiority of left/right viewpoints in the training and query sets.

\vspace{2mm}
\noindent\fbox{\parbox{0.46\textwidth}{	\textbf{Subsection conclusions}
	\vspace{-2mm}
	\begin{itemize}
		\item The query viewpoint of left/right generally leads to higher re-ID accuracy than front/back viewpoints.
		\vspace{-3mm}
\end{itemize}}}


\begin{figure}[tp]
	\begin{center}
		\includegraphics[width=0.9\linewidth]{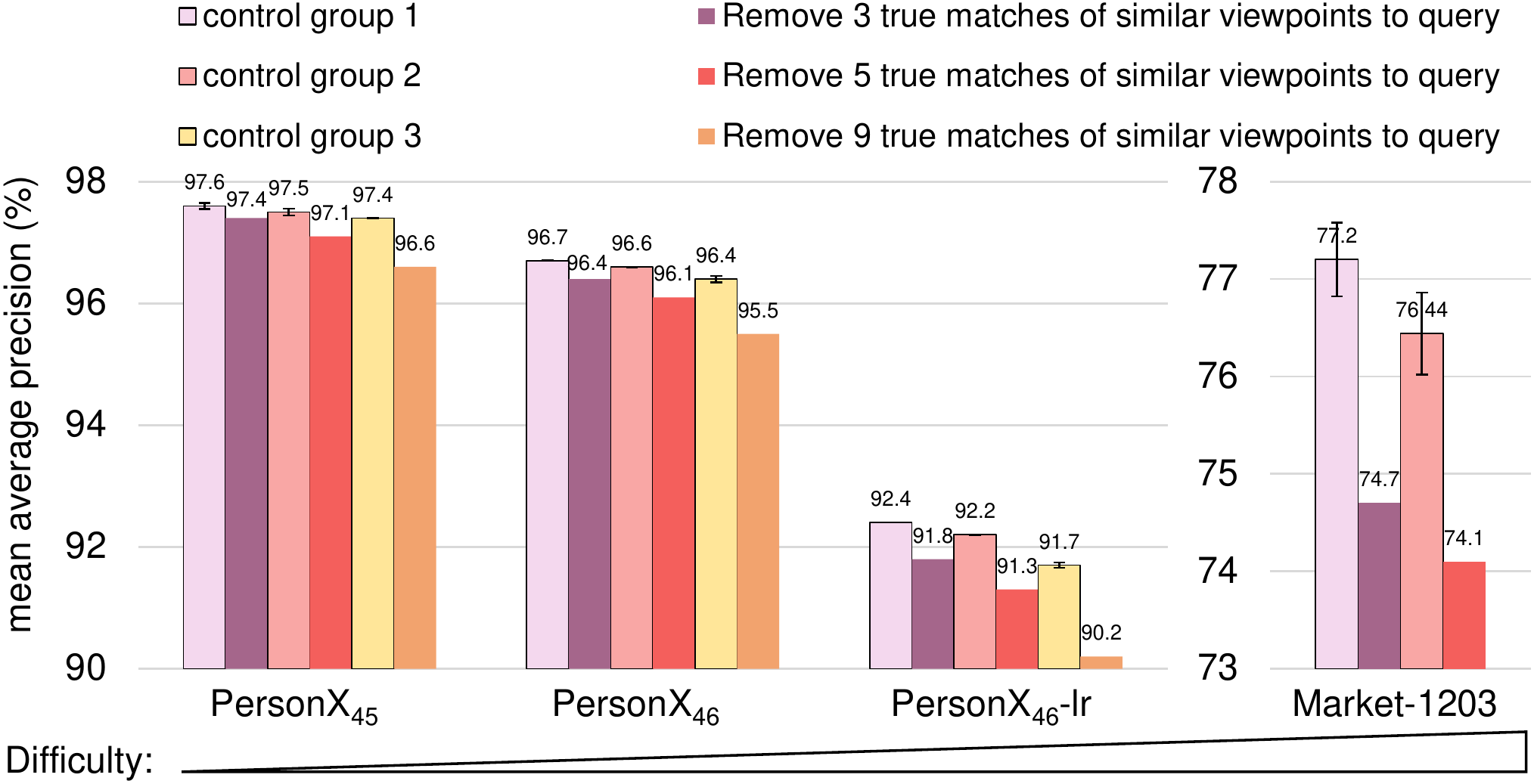}
	\end{center}
	\vspace{-3mm}
	\caption{The impact of viewpoint disparity between a query and its true matches.
            For training, we use the original training sets (balanced viewpoints). Results are reported on \textbf{A} PersonX subsets and \textbf{B} Market-1203. From PersonX$_{45}$, PersonX$_{46}$, PersonX$_{46}$-lr to Market-1203, the environmental difficulty is in increasing order. 
	}
	\label{fig:gallery}
	\vspace{-3mm}
\end{figure}

\subsection{How Do True Match Viewpoints in the Gallery Affect Retrieval?} 
Finally, we study how the gallery viewpoint distribution affects re-ID accuracy.
Specifically, we study the viewpoint disparity between the query and its true matches. 

\textbf{Experiment design.}
We denote the viewpoint of a query and its true match as $\theta_{q}$ and $\theta_{t}$, respectively.
The experimental groups for PersonX subsets are designed as below. 
\begin{itemize}
	\vspace{-2mm}
    \item Experimental group 1. The three true matches whose $\theta_{t} \in [\theta_{q}\pm 10^{\circ}]$ are removed (set as ``junk'').
	\vspace{-2mm}
    \item Control group 1. Three true matches are randomly removed from the gallery. 
    \vspace{-2mm}
    \item Experimental group 2. The nine true matches whose $\theta_{t} \in [\theta_{q}\pm 20^{\circ}]$ are removed.
    \vspace{-2mm}
    \item Control group 2. Randomly removing 5 true matches. 
    \vspace{-2mm}
    
    More difficult situations:
    \vspace{-2mm}
    \item Experimental group 3. The nine true matches whose $\theta_{t} \in [\theta_{q}\pm 40^{\circ}]$ are removed.
    \vspace{-2mm}
    \item Control group 3. Randomly removing 9 true matches. 
    \vspace{-2mm}
\end{itemize}

%
Since Market-1203 only contains eight viewpoint types, the two experimental groups remove three or five true matches that have the most similar viewpoints to the query. The corresponding control groups randomly remove the same number of true match images. 

\textbf{Result analysis.}
From Fig.~\ref{fig:gallery}, we have two observations. 

The major observation is that if true matches with similar viewpoints are not present, there will be a non-trivial performance drop. In other words, if true matches in the gallery have large viewpoint disparity with the query, the retrieval accuracy will be negatively affected.
%
%
For instance, the mAP of removing 9 true matches (experimental group) on PersonX$_{45}$ is 96.6$\%$, and there is a decrease of about 1.0$\%$ on mAP compared to the control group 3.
Consistent observation can be made on  Market-1203.
For example, compared with the control group, there is a decrease of about 3.0\% on mAP when 3 or 5 true matches with similar viewpoints to the query are removed from the gallery. 

\begin{figure}[tp]
	\begin{center}
		\includegraphics[width=\linewidth]{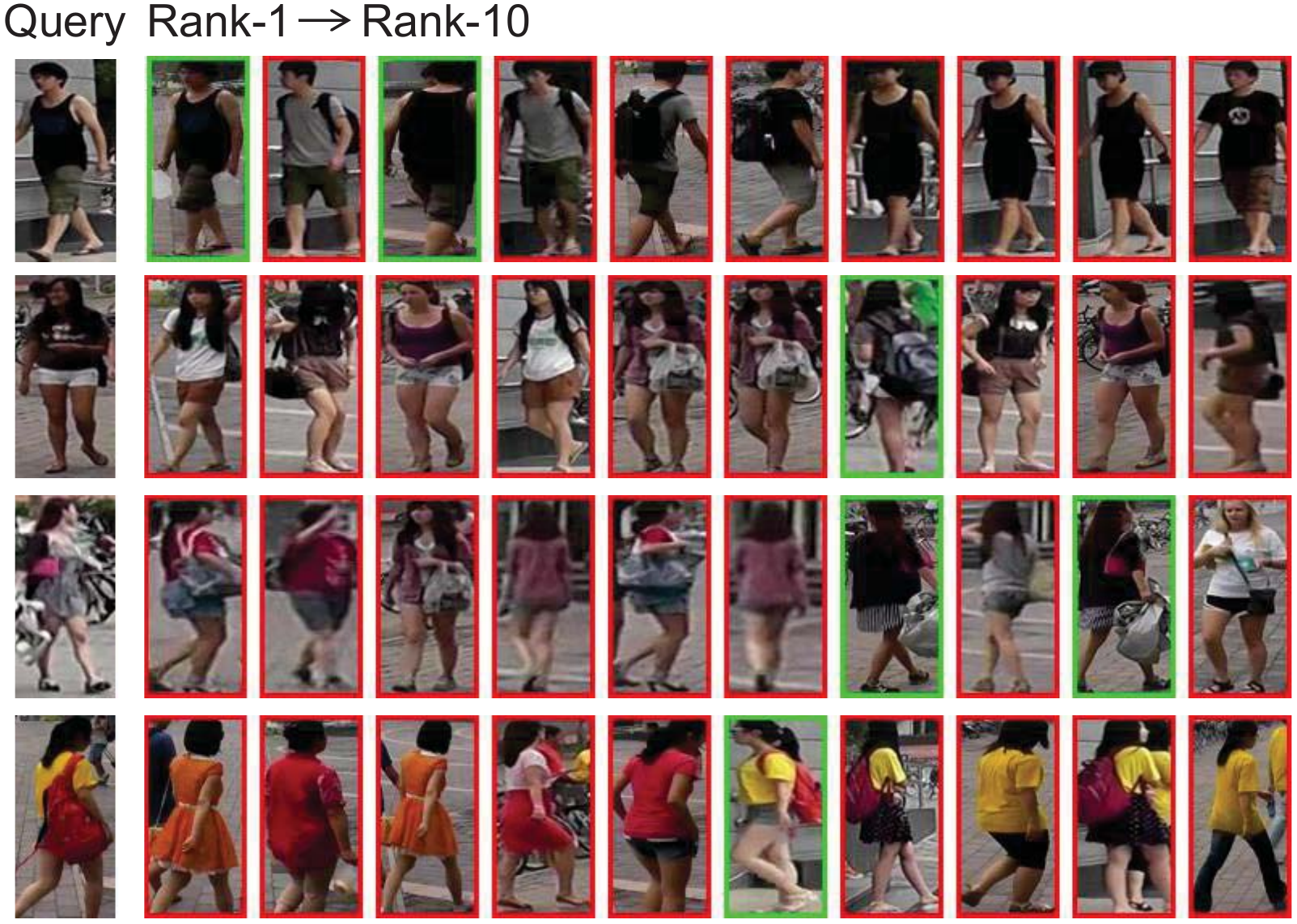}
	\end{center}
	\vspace{-3mm}
	\caption{Sample re-ID results on Market-1203. Images in the first column are queries. The retrieved images are sorted according to their similarity to the query (high to low) from left to right. The similarity is calculated by using feature extracted from the PCB model. True matches and false matches are in \textcolor{green}{green} and \textcolor{red}{red} rectangles, respectively.
	}
	\label{fig:visualization}
	\vspace{-3mm}
\end{figure}

Figure~\ref{fig:visualization} shows some re-ID results on the Market-1203 dataset.
For the first query images in Fig.~\ref{fig:visualization}, the true match is ranked to the highest position. This is because the first true match is similar to the query in both appearance and viewpoint.
After removing it, the highly ranked images are mostly false matches that have a similar viewpoint with the query.
Similarly, for other example query images that do not have true matches of similar viewpoints in gallery, the false matches of distinctive appearance (\eg different styles and colors of clothes and bags) but similar viewpoints to the query will be ranked higher than the true matches.

Moreover, the accuracy decrease caused by viewpoint disparity between a query and its true match becomes more obvious when the environment becomes more challenging.
For example, the mAP drop of the experimental groups on the PersonX$_{46}$-lr dataset is almost twice as large as the performance decline on the PersonX$_{46}$ dataset. 
%
%

%
%
%
%
%

\vspace{2mm}
\noindent	\fbox{\parbox{0.46\textwidth}{	\textbf{Subsection conclusions}
	\vspace{-2mm}
	\begin{itemize}
		\item True matches whose viewpoints are dissimilar to the query are harder to be retrieved than true matches with a similar viewpoint to the query.
		\vspace{-2mm}
		\item The above problem becomes more severe when the environment is challenging, \eg complex background, extreme illumination, and low resolution. 
		\vspace{-3mm}
\end{itemize}}}

%

\section{Conclusion}
This paper makes a step from engineering new technologies to science new discoveries. We make two contributions to the community. First, we build a synthetic data engine PersonX that can generate images under controllable cameras and environments. Subsets of PersonX are shown to be indicative of the real world. 
Second, based on PersonX, we conduct comprehensive experiments to quantitatively assess the influence of pedestrian viewpoint on person re-ID accuracy. Interesting and constructive insights are derived, \eg it is better to use a query image capturing the side view of a person. In the future, visual factors such as illumination and background will be studied with this new engine.
%
%
%
%

%
%
%
%

{\small
\bibliographystyle{ieee}
\bibliography{egbib}
}
\end{document}